\pdfoutput=1

\documentclass[11pt]{article}

\usepackage[final]{acl}

\usepackage{times}
\usepackage{latexsym}

\usepackage{tabularx}
\usepackage{booktabs}
\usepackage{hyperref}
\usepackage{amsmath}
\usepackage{graphicx}
\usepackage{caption}
\usepackage{subcaption}
\usepackage{CJKutf8}
\usepackage{algorithm}
\usepackage{algorithmic}
\usepackage{xcolor}

\definecolor{lightblue}{rgb}{0, 1, 1}
\definecolor{lightpurple}{rgb}{0.93, 0.85, 0.93}
\definecolor{darkpurple}{rgb}{0.85, 0.7, 0.85}

\usepackage[T1]{fontenc}

\usepackage[utf8]{inputenc}

\usepackage{microtype}

\usepackage{inconsolata}

\title{Empirical Study of Mutual Reinforcement Effect and Application in Few-shot Text Classification Tasks via Prompt}

\author{
  Chengguang Gan\textsuperscript{1} 
  Tatsunori Mori\textsuperscript{1} \\
  \textsuperscript{1}Yokohama National University, Japan \\
  \texttt{gan-chengguan-pw@ynu.jp, tmori@ynu.ac.jp} \\
}

\begin{document}
\maketitle
\begin{abstract}

The Mutual Reinforcement Effect (MRE) investigates the synergistic relationship between word-level and text-level classifications in text classification tasks. It posits that the performance of both classification levels can be mutually enhanced. However, this mechanism has not been adequately demonstrated or explained in prior research. To address this gap, we employ empirical experiment to observe and substantiate the MRE theory. Our experiments on 21 MRE mix datasets revealed the presence of MRE in the model and its impact. Specifically, we conducted compare experiments use fine-tune. The results of findings from comparison experiments corroborates the existence of MRE. Furthermore, we extended the application of MRE to prompt learning, utilizing word-level information as a verbalizer to bolster the model's prediction of text-level classification labels. In our final experiment, the F1-score significantly surpassed the baseline in 18 out of 21 MRE Mix datasets, further validating the notion that word-level information enhances the language model's comprehension of the text as a whole.

\end{abstract}

\section{Introduction}

\begin{figure*}[!t]
\centering
\includegraphics[width=440 pt]{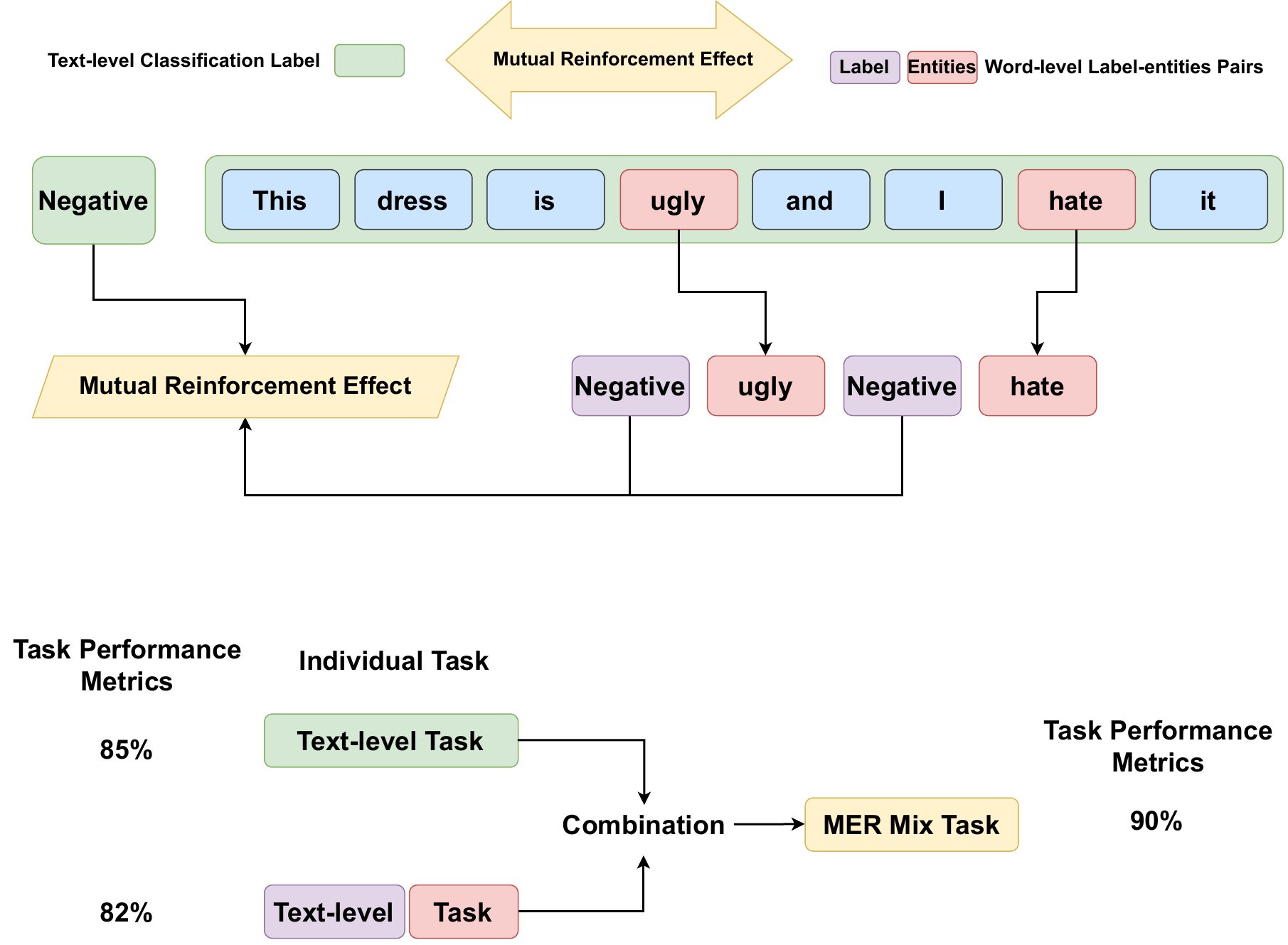}
\caption{\label{1figure}The figure illustrates the mutual reinforcement effect between text-level and word-level tasks in sentiment classification task.}

\end{figure*}

The concept of Mutual Reinforcement Effect (MRE) \citet{gan2023sentence} was initially introduced in the domains of text classification \citet{kim2014convolutional} and Named Entity Recognition (NER)\citep{nadeau2007survey}. This approach marked the first instance where a text classification task and a NER task were integrated within the same mixed dataset. Essentially, this necessitates that the model concurrently execute text-level classification and word-level information extraction (IE) for each sentence. Consequently, every text in the dataset is assigned both text-level classification labels and NER label-entity pairs. As shown in the bottom portion of Figure \ref{1figure}, Research indicates that the combined execution of these two tasks yields superior performance compared to addressing them separately.

As illustrated in the middle section of Fig. \ref{1figure}, we provide a concrete example of the MRE using a sentiment analysis task. Initially, the overall sentiment polarity of the sentence is classified as negative. When the model identifies the overall sentiment as negative, it is more likely to infer that specific negative words (e.g., "ugly," "hate") are present during the subsequent word-level classification task. Conversely, when the model detects the presence of these negative words, it will tend to classify the overall sentiment of the text as negative. This bidirectional reinforcement between sentence-level and word-level classification exemplifies the core idea of MRE.

This synergistic approach closely mirrors the human process of reading and comprehending text. Initially, humans decipher the meaning of individual words within a text. Upon grasping the significance of each word, they can then comprehend the overall message of the text. These steps correspond to the NER task and the text classification task, respectively, where the former involves understanding individual words and the latter pertains to grasping the overall meaning of the text. Furthermore, knowing the general classification of a text, such as identifying it as an advertisement, can aid in predicting the presence of specific entities like product names within the text. Similarly, in modeling, awareness of the text's overall classification can facilitate word-level information extraction. In summary, there exists a mutual reinforcement effect between text-level and word-level tasks, enhancing the performance of both.

Subsequent to its inception, MRE has been progressively adapted to a variety of tasks within the IE domain, including sentiment classification, word sentiment polarity classification, relation extraction, and event extraction. The integration of these text-level and word-level tasks has resulted in notable performance enhancements. Moreover, these tasks have been effectively addressed through the use of generative IE techniques. The evolution of models in this domain has been marked by a transition from the initial Sentence-to-label framework (SLG) \citet{gan2023sentence} employing the T5 model, to the more advanced General Information Extraction Large Language Model (GIELLM) \citep{gan2023giellm}. These models are adept at handling a mixture of text-level and word-level tasks. The advent of Large Language Models (LLMs) \citet{ouyang2022training, openai2023gpt4, touvron2023llama, touvron2023llama2} has further revolutionized the field, as the versatility of LLMs facilitates the training of a single model to efficiently manage multiple tasks, thereby simplifying the process and yielding superior performance.

Previous studies have explored ablation experiments by combining the outputs of two tasks, where the input was essentially just the text itself. This approach only confirms that the model can simultaneously improve the performance of both tasks by learning them together. However, it does not provide any insight into whether information at the word- or text-level contributes to the improvement of the other task. To address this limitation, we designed a new empirical experiment with an innovative input-output format and evaluated it using the existing 21 MRE mixed datasets. Through a straightforward and effective fine-tuning experiment, we compared the impact of incorporating word- or text-level information into the input on the performance of the corresponding output at a different level. Our findings show that word- and text-level information significantly enhances the performance of the other task, thereby providing strong evidence in support of the MRE hypothesis.

In the final, we extend the application of the MRE to few-shot learning. Specifically, we leverage word-level information as a knowledgeable verbalizer \citet{hu-etal-2022-knowledgeable} to enhance the performance of text classification tasks with limited samples. Our experimental results demonstrate a substantial improvement in classification accuracy when using word-level information as a verbalizer compared to not using it. The main contributions of this paper are summarized as follows: (1) Utilization of novelty input and output format for fine-tune ablation experiment to observe and validate the mutual reinforcement effect. (2) Application of the MRE concept to few-shot learning in text classification tasks, resulting in a significant enhancement effect.

\section{Related Work}

\textbf{Mutual Reinforcement Effect}. In traditional multi-task \citet{zhang2021survey, liu2016recurrent, collobert2008unified, radford2019language} approaches that simply amalgamate disparate tasks, MRE integrates two inherently related tasks at the word and text levels. By curating a novel dataset, enhance the performance of both tasks concurrently. Initially, the Sentence Classification and Named Entity Recognition (SCNM) \citet{gan2023sentence} task was developed by combining sentence classification and named entity recognition. To address this integrated task, the Sentence-to-Label Generation (SLG) framework was proposed based on the T5 model. Subsequently, a new task was formulated by merging text sentiment classification with word-level sentiment polarity classification (SCPOS) \citep{gan2023usa}, marking the first application of large language models (LLMs) to train on SCPOS tasks. Following this, text classification was integrated with relationship and event extraction (TCREE) \citep{gan2023giellm}. A comprehensive General Information Extraction Language Model (GIELLM) was then constructed by incorporating the SCNM and SCPOS tasks. Finally, open-domain NER mixed datasets (TCONER) \citep{gan2024mmm} were created and extended to Chinese and English using a dataset translation framework, thereby enhancing the multilingual capability of the original Japanese MRE mixed datasets.

The effectiveness of the proposed Mutual Reinforcement Effect (MRE) is validated through a series of ablation studies, where models were trained on each task independently and then jointly on combined tasks. Comparative analysis on test sets reveals that the joint training approach consistently outperforms single-task models, demonstrating the synergistic advantages of the MRE framework.

\textbf{Prompt in Information Extraction}. The prompt-based approach was initially applied to text and sentence sentiment classification \citep{schick2021exploiting, zhong2021factual, li2021prefix, shin2020autoprompt}. It achieves few-shot learning by aligning the downstream task format with the pre-training objective of the pre-trained language model (PLM), specifically the Masked Language Modeling (MLM) task \citep{devlin-etal-2019-bert}. The original prompt method relies on a basic prompt template, wherein prompt questions are embedded within the model’s input sequence alongside the text to be classified. The model then outputs the corresponding predicted classification labels.

Subsequent research introduced the concept of prompt-based label generation, known as the Knowledgeable Verbalizer (KV) \citep{hu2022knowledgeable}. This method utilizes words associated with classification labels to aid the model in predicting the appropriate labels, thereby enriching the representation and prediction capability of the model.

In summary, the MRE is highly suitable for integration with the Knowledgeable Verbalizer. MRE not only enhances task performance but also serves as evidence that word-level information can effectively support text-level understanding.

\begin{figure*}[!t]
\centering
\includegraphics[width=440 pt]{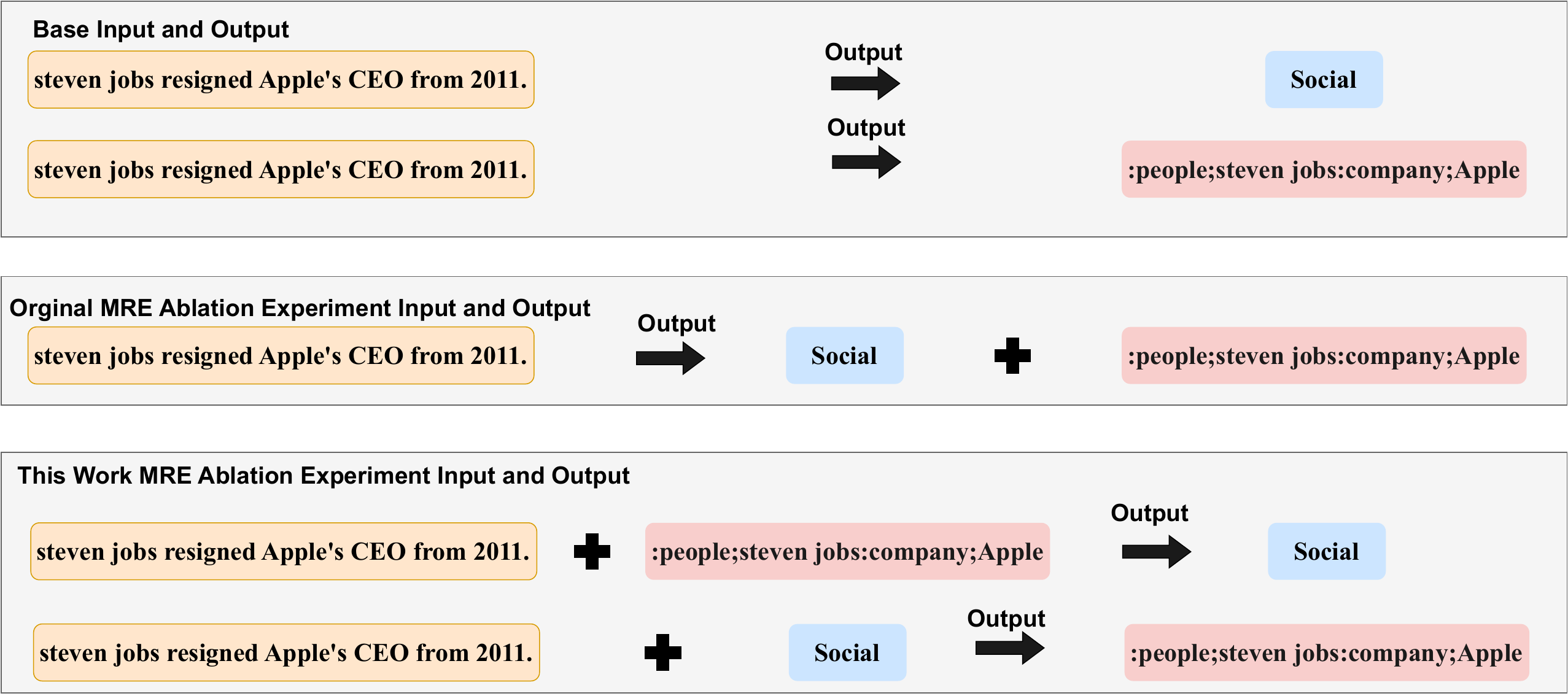}
\caption{\label{2figure}The figure shows the inputs and outputs of the traditional ablation experiment for the MRE task and the new empirical MRE experiment proposed in this work.}

\end{figure*}

\section{Empirical Experiment of Mutual Reinforcement Effect}

The three format of fine-tuned language models used for ablation experiments are shown in Figure \ref{2figure}. The sentence on the left represents the input, with the plus sign indicating the addition of Word-level Information (WLI. i.e. Word-level Task) or Text-level Information (TLI. i.e. Text-level Task), which are appended to the sentence to form the full input. The arrows represent the output produced by language model. The distinctions between the models are clearly illustrated.

First, the top model in Figure \ref{2figure} shows the input-output format for the traditional IE task, where language models are fine-tuned on a basic input sentence. The model then outputs either classified labels or extracted label-entity pairs. This approach treats the two tasks—word-level label extraction and text-level classification—independently, with no shared information between them.

In contrast, the middle section of Figure \ref{2figure} illustrates the input-output format for the original MRE task. While the input remains a single sentence, the model is expected to output both word-level label-entity pairs and text-level classification labels simultaneously. Thus, during MRE fine-tuning, the model learns to capture both levels of information, integrating the two tasks.

Finally, the bottom section of Figure \ref{2figure} presents the input-output format of our proposed ablation experiment designed to validate MRE. Unlike the previous two formats, this approach aims to verify the existence of shared knowledge between word-level and text-level tasks. Specifically, we introduce WLI and TLI to both levels of tasks to assess whether enhancing one task also improves the other. For example, by adding word-level label-entity pairs to the input text and asking the model to output the text-level classification label, we can evaluate whether the additional word-level information assists in text classification. Similarly, if adding text-level information to the input improves the extraction of word-level label-entity pairs, it suggests the presence of an MRE between the two tasks.

\begin{figure}[!h]
\centering
\includegraphics[width=219 pt]{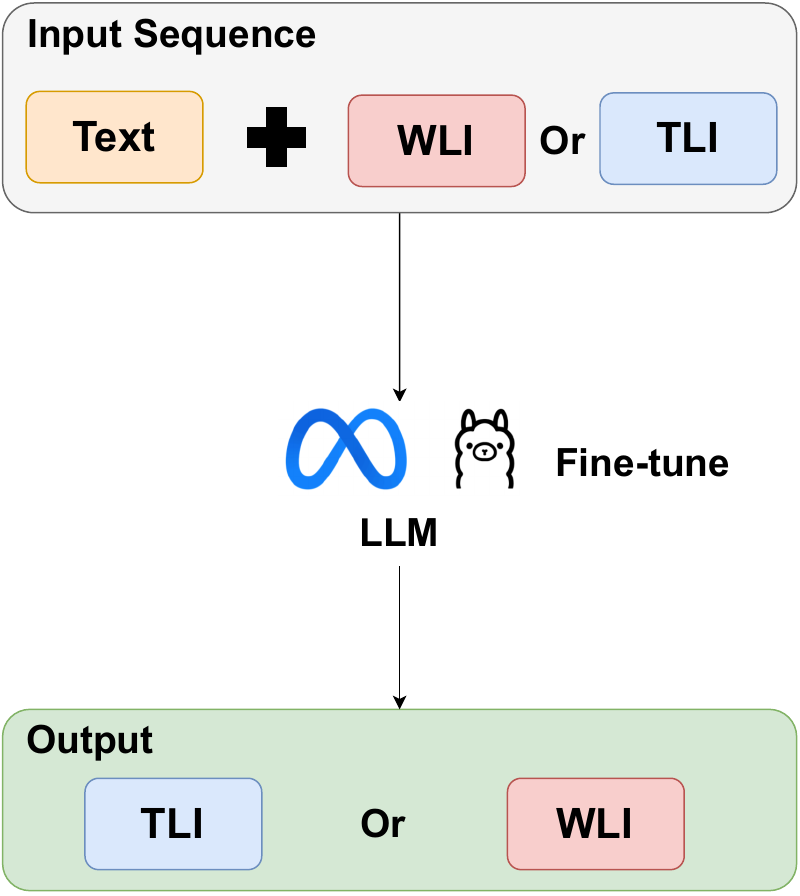}
\caption{\label{3figure}The figure illustrates the flow of an empirical MRE experiment using the new approach.}

\end{figure}

As showed in Figure \ref{3figure}, the LLM is fine-tuned with all parameters using revised input and output formats. The input sequence is directly concatenated with either WLI or TLI, while the output consists solely of TLI or WLI. No additional instruction templates or prompt words were incorporated in this process. We deliberately chose to concatenate the text with WLI or TLI without extra modifications to minimize the potential influence of extraneous words or sentences on the model’s output, which could affect the accuracy of our comparative experiments. By using only this basic spliced input and raw output, we aim to investigate whether tasks at one level facilitate tasks at another, while controlling for other confounding factors.

To test this hypothesis, we conducted ablation experiments on 21 sub-datasets of Multilingual MRE Mix (MMM) datasets\citep{gan2024mmm}. The results were analyzed to further deepen our understanding of MRE and its implications.

\begin{figure*}[!t]
\centering
\includegraphics[width=440 pt]{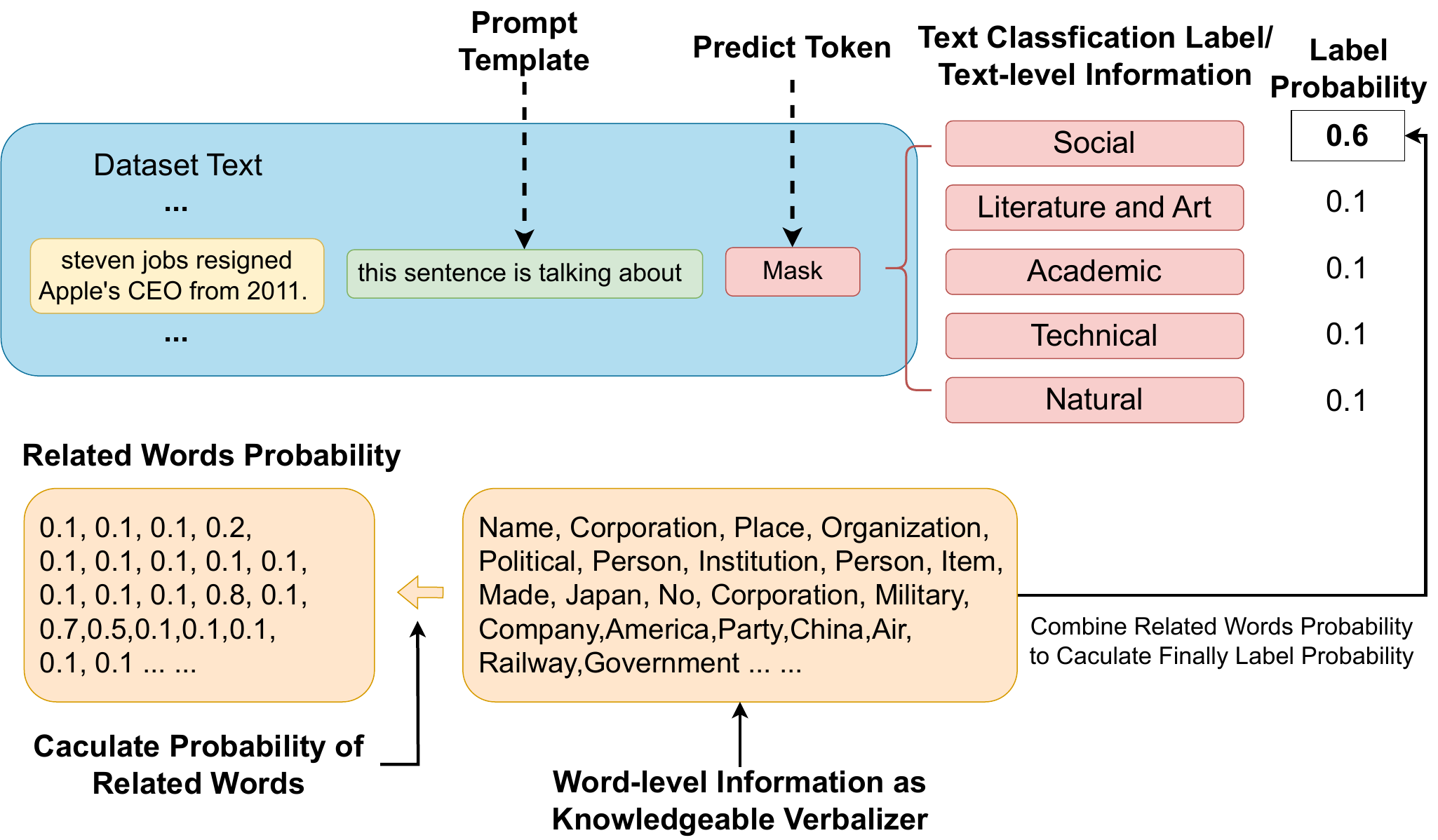}
\caption{\label{4figure}The figure demonstrates how word-level information is utilized as a Knowledgeable Verbalizer to assist in text-level classification tasks. Additionally, it provides a detailed explanation of the functioning of the Knowledgeable Verbalizer.}

\end{figure*}

\section{Word-level Information as Knowledgeable Verbalizer}

To enhance the application of the MRE approach in real-world contexts, we have selected the few-shot learning task for text classification as our experimental setup. In MRE, word-level information plays a crucial role in text-level classification. Hence, we utilize the high-frequency words from word-level information as knowledgeable verbalizers (KV) \citet{hu-etal-2022-knowledgeable} to examine their impact on the performance of the text classification task.

The entire process of prompt learning is illustrated in Figure \ref{3figure}. Initially, a target classification text is provided, followed by the inclusion of a prompt template to guide the model in predicting the label at the designated mask position. Our sample dataset comprises five labels. We employ the top 100 words from the word-level information as the knowledgeable verbalizer, meaning that each of the five categories has 100 high-frequency words selected from the word-level information. When calculating the actual probability of a label, the model computes the probability of all these 500 words and then aggregates the total probability based on the respective broad classification. Ultimately, we obtain five probabilities that integrate the individual verbalizers. The label with the highest probability is chosen as the final predicted label.

In conclusion, this outlines the detailed principle behind the KV. In the original experimental setup, label-related high-frequency words were sourced directly from a relation word search website, where commonly used vocabulary was analyzed to identify relevant terms. While these words may be highly pertinent across a wide range of web texts, not all of them are necessarily associated with the labels of a specific dataset. As a result, some of these words may not only fail to enhance label prediction but could potentially introduce negative effects. This highlights the suitability of the WLI component from the MRE-mixed dataset as a replacement for the KV. Furthermore, if the performance of the WLI-based KV surpasses that of the original baseline KV, it would support the argument that WLI contributes positively to label prediction in text classification tasks. This, in turn, would verify the presence of the MRE.

\section{Experiment Setup}

\begin{table*}[!t]
\centering
\begin{tabular}{lll}
\hline
 Datasets & \textbf{Text-level} & \textbf{Word-level}    \\
\hline
 SCNM & Society, Literature,  & people, corporations, political \\
  & Academia, Technology, &   organizations, other organizations, \\
& Nature  &   places, facilities, products, and events     \\
\hline
 SCPOS:RW& positive, negative & positive, neutral, negative \\
 \hline
 SCPOS:N& positive, negative & positive, neutral, negative   \\
 \hline
 SCPOS:Adj& positive, negative & positive, negative  \\
 \hline
 SCPOS:N \& Adj& positive, negative & positive, neutral, negative  \\
 \hline
 TCREE & sports, film, women,  & affiliation, occupation, starring, director,  \\
  & IT, advertising  &  age, product, goods, performances, wins,    \\
  &   &   broadcasts, public appearances, launches,  \\
  &   &   retirements  \\
  \hline
  TCONER & Entertainment, Politics & date, location, organization   \\
         & Medical, Health, education  & Title, Person, City  \\
         & Tech, Healthcare, News  & Law, Number, Concept    \\
         & finance, Biolog, etc.  &  TV Show, Object, etc.  \\
\hline
\end{tabular}
\caption{\label{table1mixdataset}
The table presents seven distinct types of MRE mixed datasets, each available in Chinese, English, and Japanese, resulting in a total of 21 sub-datasets. Among them, the TCONER dataset corresponds to an open-domain dataset, where only a subset of the labels is provided, rather than a comprehensive list of all possible labels. (SCNM: Sentence Classification and Named Entity Recognition Mix Dataset. SCPOS: Sentiment Classification and Part-of-Speech Dataset. RW: Relation Word. N: Noun. Adj: Adjective. N \& Adj: Nous and Adjective. TCREE: Text Classification and Relation \& Event Extraction Dataset. TCONER: Open-domain Text Classification and NER mix dataset)}
\end{table*}

For the fine-tuning experiment, the entire training set was utilized to fully parameterize the fine-tuned LLMs. Subsequently, 1,000 samples were randomly selected from the test set three times, and the results from these three trials were averaged to produce the final performance score. The evaluation metric employed was the F1 score.

The hyperparameters for training were configured as follows: the number of training epochs was set to 3, and the learning rate was initialized at 1e-5. The AdamW optimizer was used, with 100 warm-up steps. Training was conducted on three RTX A6000 Ada GPUs, each with 48 GB of memory. To optimize GPU memory usage, BF16 precision was applied during training, and FP16 precision was employed for inference.

Second, for the experiments involving the knowledgeable verbalizer, we utilized the OpenPrompt\citet{ding2021openprompt}\footnote{https://github.com/thunlp/OpenPrompt} framework to efficiently set up the experimental environment. All datasets were divided into training and test sets. From the training set, we randomly selected 20 samples per category, based on the label types, to form the prompt experiment’s training subset. Each experiment was trained for 2 epochs, with all other hyperparameters—such as the learning rate—kept consistent across experiments. The only variation lay in the construction method of the KV.

For the KVs based on the original approach, we leveraged ChatGPT-4o\footnote{https://chatgpt.com/} to generate the top 100 most relevant words for each label. In contrast, for KVs constructed using the WLI-based method, we developed a custom processing script. The script segmented all words from the WLI section of each dataset, identified high-frequency terms, and used them to construct the WLI-based KVs.

\subsection{Datasets}

In the selection of datasets, we focused on Multilingual MRE Mix (MMM) datasets\footnote{https://huggingface.co/datasets/ganchengguang/
MMM-datasets-Testset and https://huggingface.co/datasets/
ganchengguang/MMM-dataset-Trainset} due to their availability. We tested 21 such datasets, the statistics of which are presented in Table \ref{table1mixdataset}. These datasets include SCNM (Sentence Classification and NER mix dataset), SCPOS: RW/Adj\&N/Adj/N (Sentiment Classification and Part-of-Speech dataset: Related Word/Adjective\&Noun/ Adjective/Noun), TCREE (Text Classification and Relation Event Extraction dataset). and TCONER (Open-domain Text Classification and NER mix dataset). These datasets cover a wide range of tasks, encompassing nearly all divisions of IE. Moreover, there is a correlation between the word-level labels and text-level labels within these datasets. For instance, the sentiment polarity of words is related to the sentiment polarity of the text. It is important to note that the TCONER dataset, being open-domain, contains labels with inherent uncertainty. Therefore, we excluded TCONER from the KV application experiments to ensure reliable evaluation.

\subsection{Model}

\begin{table*}[!t]
\centering
\begin{tabular}{lcccccc}
\hline
 English & \textbf{SCNM} & \textbf{SCPOS:RW} & \textbf{SCPOS:adj\&n} & \textbf{SCPOS:adj}& \textbf{SCPOS:n}  & \textbf{TCREE}  \\
\hline
w/o TLI & 80.97 & 48.79 & \textbf{33.29} & 56.04 & \textbf{28.79} & 16.43 \\
with TLI & \textbf{81.28} & \textbf{48.99} & 32.42 & \textbf{56.75} & 27.71 & \textbf{18.43}\\
w/o WLI & 82.40 & 72.41 & 77.27 & 73.73 & 77.07 & 82.23 \\
with WLI & \textbf{83.90} & \textbf{73.15} & \textbf{77.60} & \textbf{75.70} & \textbf{77.73} & \textbf{83.33} \\
\hline
 Chinese & \textbf{SCNM} & \textbf{SCPOS:RW} & \textbf{SCPOS:adj\&n} & \textbf{SCPOS:adj} & \textbf{SCPOS:n}  & \textbf{TCREE}  \\
\hline
w/o TLI & \textbf{73.35} & \textbf{44.36} & 28.67 & 9.68 & 29.06 & 55.10 \\
with TLI & 72.81 & 43.30 & \textbf{29.17} & \textbf{9.73} & \textbf{29.34} & \textbf{56.31} \\
w/o WLI & 83.17 & 89.07 & 91.03 & \textbf{93.67} & 91.80 & 93.64 \\
with WLI & \textbf{83.93} & \textbf{90.95} &  \textbf{92.37} & 92.07 & \textbf{93.63} & \textbf{94.85} \\
\hline
 Japanese & \textbf{SCNM} & \textbf{SCPOS:RW} & \textbf{SCPOS:adj\&n} & \textbf{SCPOS:adj}& \textbf{SCPOS:n}  & \textbf{TCREE}  \\
\hline
w/o TLI & 87.92 & 69.47 & 63.80 & 50.70 & \textbf{67.23} & 80.87 \\
with TLI & \textbf{88.22} & \textbf{69.92} & \textbf{63.89} & \textbf{51.03} & 66.24 & \textbf{81.37} \\
w/o WLI & 83.60 & 87.10 & 88.13 & 87.93 & 88.37 & \textbf{94.86} \\
with WLI & \textbf{85.87} & \textbf{89.50} & \textbf{89.17} & \textbf{89.90} & \textbf{90.57} & 94.46 \\
\hline
 \textbf{TCONER} & English &  & Chinese &  & Japanese &   \\
\hline
w/o TLI & \textbf{20.22} &  & 17.28 &  & 13.19 &  \\
with TLI & 19.85 &  & \textbf{17.82} &  & \textbf{13.39} & \\
w/o WLI & \textbf{36.50} &  & \textbf{44.07} &  & 38.97 &  \\
with WLI & 35.53 &  & 43.33 &  & \textbf{43.30} &  \\

\hline
\end{tabular}
\caption{\label{table2MREresults}
The results of text-level information (TLI) and word-level information (WLI) comparison experiments.
}

\end{table*}

For the empirical experiments on fine-tuning, we selected the LLaMA3-8B\footnote{https://ai.meta.com/blog/meta-llama-3/} model\footnote{https://huggingface.co/meta-llama/Meta-Llama-3-8B} as the base model to perform a series of fine-tuning and inference tasks. We opted not to use the LLaMA3-8B-Instruct version because it is more tailored for question-answering tasks, with prompts structured as instructions. Through a comparative analysis of LLaMA3-8B and its instruct-tuned counterpart, we observed that the base LLaMA3-8B model achieved better performance on fundamental IE tasks. Therefore, we decided to use LLaMA3-8B as the foundation for our experiments.

For the WLI as KV application comparison experiments, we employed the T5-base \citet{raffel2020exploring} model as the base model. Specifically, for the English portion of the MMM dataset, we used the original Google T5-base\footnote{https://huggingface.co/google-t5/t5-base}. For the Chinese section, we selected the Mengzi-T5-base\footnote{https://huggingface.co/Langboat/mengzi-t5-base}, which is optimized for Chinese tasks. Lastly, for the Japanese part of the MMM dataset, we utilized T5-base-Japanese\footnote{https://huggingface.co/sonoisa/t5-base-japanese}.

\section{Results}

\subsection{Empirical Experiment of Mutual Reinforcement Effect}

As shown in Table \ref{table2MREresults}, the results of our test set consist of 84 LLaMA3-8B models fine-tuned on 21 datasets in four distinct formats. The first column lists the names of seven core datasets, with the seventh dataset, TCONER, discussed separately at the bottom of the table due to its unique characteristics. The four formats are divided as follows: the first two involve TLI (text-level information), where it is either excluded or included in the input sequence, with the output mapped to WLI (word-level information). The remaining two formats address WLI, where it is similarly either excluded or included in the input sequence, with the output corresponding to WLI.

\begin{table*}[!t]
\centering
\begin{tabular}{lcccccc}
\hline
 English & \textbf{SCNM} & \textbf{SCPOS:RW} & \textbf{SCPOS:adj\&n} & \textbf{SCPOS:adj}& \textbf{SCPOS:n}  & \textbf{TCREE}  \\
\hline
Origin KV & 62.95 & 80.42 & 80.40 & 78.87 & 81.95 & \textbf{86.52} \\
WLI KV & \textbf{63.24} & \textbf{83.99} & \textbf{87.40} & \textbf{87.37} & \textbf{88.70} & 85.82 \\
\hline
 Chinese & \textbf{SCNM} & \textbf{SCPOS:RW} & \textbf{SCPOS:adj\&n} & \textbf{SCPOS:adj} & \textbf{SCPOS:n}  & \textbf{TCREE}  \\
\hline
Origin KV & 67.38 & 78.37 & \textbf{91.90} & 84.48 & 84.45 & 93.04 \\
WLI KV & \textbf{71.96} & \textbf{87.97} & 82.92 & \textbf{88.38} & \textbf{87.23} & \textbf{93.95} \\
\hline
 Japanese & \textbf{SCNM} & \textbf{SCPOS:RW} & \textbf{SCPOS:adj\&n} & \textbf{SCPOS:adj}& \textbf{SCPOS:n}  & \textbf{TCREE}  \\
\hline
Origin KV & 73.26 & 30.20 & 67.23 & 73.71 & 73.71 & 73.11 \\
WLI KV & \textbf{73.91} & \textbf{52.90} & \textbf{81.74} & \textbf{85.67} & \textbf{88.31} & \textbf{77.24} \\

\hline
\end{tabular}
\caption{\label{table3MREKVresults}
The results of word-level information (WLI) as knowledgeable verbalizer experiments. Compare with original KV construction method. Evaluation task is text classification task.
}

\end{table*}

From the results in Table \ref{table2MREresults}, we observe that for the first six fixed-label datasets, models trained with the inclusion of additional information consistently outperform those trained without it. These findings strongly support the MRE hypothesis, demonstrating that mutual reinforcement exists between word-level and text-level classification tasks. A well-balanced combination of both classification levels enhances the LLMs ability to understand and perform across tasks. Specifically, comprehension of one task level (e.g., text-level) facilitates and strengthens the understanding of the other (e.g., word-level).

This insight not only advances our understanding of how LLMs tackle natural language tasks but also reflects a broader principle underlying human cognition: the mutual reinforcement between different levels of text comprehension mirrors how humans naturally process and understand language.

As illustrated by the results of the open-domain text classification and NER tasks at the bottom of Table \ref{table2MREresults}, approximately half of the outcomes do not surpass those achieved by the model trained without Level Information. We attribute this to the nature of certain open-domain datasets, which contain multiple labels; in such cases, not all WLI contributes positively to TLI. The presence of these uncorrelated WLIs and TLIs leads to a decline in overall performance. However, in the Chinese and Japanese TCONER datasets, we observe improved results after incorporating Level Information. This improvement suggests that the MRE is more effective in languages based on Chinese characters, in contrast to those that use alphabetic writing systems, such as English.

\subsection{Word-level Information as Knowledgeable Verbalizer}

The next result involves the use of WLI as the relevant word for constructing KVs. We compare the performance of KVs constructed using the original method with those built using WLI in a text classification task. Since KV construction requires a fixed label structure, the open-domain TCONER dataset, which has an unfixed label schema, was excluded from this experiment.

As shown in Table \ref{table3MREKVresults}, across 18 sub-datasets in English, Chinese, and Japanese, the WLI-based KVs achieved the highest performance in 16 datasets. Moreover, for most sentiment classification datasets, KVs constructed with WLI significantly outperformed those generated by the original method in terms of F1 scores. These results not only demonstrate the effectiveness of WLI in enhancing general text classification tasks but also highlight its particular value in sentiment classification. This is likely because sentiment classification heavily relies on correctly identifying the sentiment polarity of individual words within the text, which aligns with WLI’s strengths.

\section{Conclusion}

In this study, we propose novel input and output schemes to rigorously test and validate the Mutual Reinforcement Effect (MRE) hypothesis. Extensive empirical experiments are conducted using LLMs, with the results confirming both the existence and validity of MRE. Furthermore, we apply MRE to practical text classification tasks, demonstrating its effectiveness. Specifically, the knowledgeable Verbalizer (KV) structure constructed through the Word-Level Information (WLI) approach outperforms the original method in text classification tasks. We believe these findings will serve as a valuable reference for future researchers, facilitating further exploration and application of MRE.

\section{Limitation}

Due to constraints in resources and time, this study focused solely on experiments with the LLAMA3-8b model. In future work, we aim to extend our experiments to additional LLMs for broader validation. Furthermore, in this application of the MRE theory, we applied the WLI to TLI. However, the reverse interaction applying TLI to WLI was not explored. We plan to address these limitations and explore these directions in future research.


\bibliography{custom}

\appendix



\end{document}